%% file: eacl23.tex
\pdfoutput=1

\documentclass[11pt]{article}

\usepackage[]{EACL23}

\usepackage{times}
\usepackage{latexsym}
\usepackage{amsmath}
\usepackage{amsfonts}
\usepackage{cleveref}
\usepackage{subfiles}
\usepackage{graphicx}
\usepackage{multirow}
\usepackage{booktabs}
\usepackage[T1]{fontenc}

\usepackage[utf8]{inputenc}
\usepackage[ruled, vlined, linesnumbered]{algorithm2e}

\usepackage{microtype}
\usepackage{bibunits}

\DeclareMathOperator*{\argmax}{arg\,max}

\title{Step by Step Loss Goes Very Far: \\ Multi-Step Quantization for Adversarial Text Attacks}

\author{Piotr Gaiński \\
  Jagiellonian University \\
  Ardigen \\
  piotr.gainski@doctoral.uj.edu.pl \\\And
  Klaudia Bałazy \\
  Jagiellonian University \\
  klaudia.balazy@doctoral.uj.edu.pl \\}

\begin{document}
\maketitle

\begin{abstract}
We propose a novel gradient-based attack against transformer-based language models that searches for an adversarial example in a continuous space of token probabilities. Our algorithm mitigates the gap between adversarial loss for continuous and discrete text representations by performing multi-step quantization in a quantization-compensation loop. Experiments show that our method significantly outperforms other approaches on various natural language processing (NLP) tasks.

\end{abstract}

\urlstyle{same}

\section{Introduction}
\subfile{content/introduction}
\section{Related Work}
\subfile{content/background}

\section{MANGO}
\label{sec:mango}
\subfile{content/mango}
\subfile{content/tables/all_results}

\section{Experiments}
\label{sec:experiments}
\subfile{content/experiments}

\section{Ablation Study}
\label{app:entropy}
\subfile{content/ablations}

\section{Visualization of Quantization Gap}
\label{app:losses}
\subfile{content/visualization}



\section{Conclusion}
\subfile{content/conclusion}

\section*{Limitations}
\subfile{content/limitations}

\section*{Acknowledgements}

The work of Klaudia Bałazy was carried out within the research project "Bio-inspired artificial neural network" (grant no. POIR.04.04.00-00-14DE/18-00) within the Team-Net program of the Foundation for Polish Science co-financed by the European Union under the European Regional Development Fund. Piotr Gaiński and Klaudia Bałazy are affiliated with Doctoral School of Exact and Natural Sciences at the Jagiellonian University.


\bibliography{eacl23}


\appendix
\subfile{content/appendix}

\end{document}

%% file: content/introduction.tex
Deep neural networks achieve impressive results, but their vulnerability to adversarial attacks causes major security threats and is a concern when interpreting or explaining model predictions.

In computer vision, the most successful attack methods use gradient-based optimization techniques \cite{im1,im2}. They minimize adversarial loss function that encourages the prediction error and imperceptibility of a generated example.

Development of optimization-based attacks in NLP is much more challenging due to the discrete nature of text. Recent methods \cite{gbda,bridge_the_gap} overcome this limitation by performing a gradient descent in the continuous space of token representations and eventually quantizing them into discrete text.

A quantization of a token can significantly change its embedding and cause an undesired change of the loss value, degrading the adversarial example. To our knowledge, all existing optimization-based NLP attacks quantize all tokens in a text at once, which creates a considerable gap between adversarial loss for continuous and discrete text representations.

In this paper, we propose MANGO\footnote{Code available at \url{github.com/gmum/MANGO}.} (Multi-step quANtization Gradient-based adversarial Optimizer): a novel optimization-based attack against Transformer \cite{transformer} language models that mitigates the aforementioned gap by performing multi-step quantization in a quantization-compensation loop.
MANGO quantizes continuous token representations one by one and reoptimizes the adversarial example after each quantization to compensate undesired degradation of adversarial loss value. 
The construction of MANGO introduces interesting problems that are addressed in \Cref{sec:mango}. MANGO achieves superior performance in various NLP tasks, outperforming recent white-box (optimization-based) and black-box attacks.

%% file: content/background.tex
Adversarial attacks can be roughly divided into two categories: white-box attacks that have access to the internal model's states (e.g. gradient) and more common black-box attacks that only know outputs of the model. In our paper, we focus on a white-box version of our MANGO attack. In \Cref{app:black_mango}, we develop a version of MANGO that can be used in the loosened black-box setting.

\paragraph{Black-Box Methods} Most black-box NLP attacks define a space of character or word replacements and heuristically search it for an adversarial example \cite{search}. The search space is limited with semantic ad hoc constraints (e.g. limiting edit distance or restricting possible replacements to synonyms) to preserve the attack's imperceptibility. Such constraints disallow some specific perturbations (e.g. replacing a word with its antagonist even if the semantics is preserved in the context of other perturbations) and tend to generate semantically incorrect examples~\cite{reeval}.

\paragraph{White-Box Methods}
Many white-box methods use gradients to guide a heuristic search in a space of text perturbations \cite{hotflip,cheng-etal-2019-robust,texttricker}. 
Recent methods take a step further and perform gradient descent optimization. They aim to find an example that minimizes the adversarial loss function, which encourages the prediction error and the imperceptibility of the attack. Because the similarity and fluency of an example are controlled by a powerful external model used in the loss, optimization-based methods do not require hand-crafted semantic constraints, making them more flexible than black-box ones.

Adapting gradient descent in NLP attacks is a challenging problem due to the discrete nature of the optimized text. \citet{bridge_the_gap} overcome this issue by performing optimization in the continuous space of token embeddings and replacing each token with a possibly new token, which embedding is the closest to the optimized one. An alternative approach is the GBDA method~\cite{gbda} that optimizes a continuous distribution of stochastic one-hot vectors and repeatedly samples adversarial examples from the optimized distribution until it fools the attacked model. 

\paragraph{Quantization} Both methods mentioned above quantize all continuous representations of tokens to a text at once. Quantization of a single token may significantly change its embedding and cause an undesirable change of adversarial loss value. When quantizing all tokens at once, the changes accumulate to a considerable gap between adversarial loss for continuous and discrete text representations (see~\Cref{app:losses}). Our MANGO mitigates this gap.

%% file: content/mango.tex
This section describes our MANGO method. Unlike other optimization-based methods that quantize all token representations at once, MANGO constitutes an entirely new algorithm that quantizes a token and compensates for the resulting change in an adversarial loss value in a step-by-step manner. The construction of MANGO introduces interesting problems that are addressed in the \textbf{Optimization}, \textbf{Vector Selection} and \textbf{Candidates Selection} paragraphs and are further evaluated in \Cref{app:entropy}.

\paragraph{Continuous Token Representation}
The first learnable layer of Transformer takes as input a sequence of tokens $x=(t_1, ..., t_n)$, where $t_i \in 2^{|V|} $ has a single non-zero binary value at index $k$ indicating that it represents the $k$-th token in vocabulary $V$.

Similarly to \citet{gbda}, we relax the input sequence $x$ and replace one-hot encodings $t_i$ with probability vectors $\pi_i$. Because the first learnable Transformer layer is a simple linear layer, it can take probability vectors as input without any modification.

A probability vector $\pi_i$ constitutes probability distribution over tokens~from~$V$. 

In the embedding layer, the Transformer embeds probability vectors with the function $e$:

\begin{equation}
    e(\pi_i) = \sum_{j=1}^{|V|} (\pi_i)_j E_j,
\end{equation}
where $E_j$ is the embedding vector of the $j$-th token. If $\pi_i$ is quantized, meaning it is a one-hot vector representing some token $k$, function $e$ simply looks up the $k$-th embedding: ${e(\pi_i)=E_k}$. In MANGO, $\pi_i$ is a probabilistic vector, and its embedding $e(\pi_i)$ is a mixture of embeddings of all tokens weighted by their probabilities $\pi_i$. We parameterize $\pi_i$ with logits $\Theta_i$ and a standard softmax function $\sigma$, so that $\pi_i=\sigma(\Theta_i)$ and $x=\sigma(\Theta)$ for $\Theta=(\Theta_1,...,\Theta_n)$.

\paragraph{Loss function}
Let $m: X \rightarrow \mathbb{R}^{|Y|}$ be a classifier that outputs logit vectors and properly predicts a label $y\in Y$ for some datapoint $x\in X$, meaning that $\argmax_k m(x)_k=y$. An adversarial example is a sample $x' \in X$ that is imperceptible (according to specified criteria) from $x$  but changes the output of the model. In an optimization-based setting, searching for an adversarial example is usually defined as a minimization of an adversarial loss function.

Following \citet{gbda}, we compose our adversarial loss $\mathcal{L}$ as a combination of margin loss $l_m$, fluency loss $l_f$, and similarity loss $l_s$:
\begin{equation}
\label{eq:loss}
    \mathcal{L}(x') = l_m(m, x', y) + \lambda_f l_f(g, x') + \lambda_s l_s(g, x', x),
\end{equation}
where $\lambda_f$ and $\lambda_s$ are the coefficients used to balance the losses and $g$ is a reference model. 

Margin loss $l_m$ encourages model $m$ to missclassify $x'$ by a margin $\kappa$:
\begin{equation*}
    l_m(m, x', y) = \max(m(x')_y - \max_{k \neq y} m(x')_k + \kappa, 0).
\end{equation*}

Fluency loss $l_f$ promotes $x'$ with a high probability of being generated by a causal language model $g$ that predicts the next token distribution:
\begin{equation*}
    l_f(g, x') = -\sum_{i=1}^n\sum_{j=1}^{|V|} (\pi_i)_jg(\pi_1, ..., \pi_{i-1})_j.
\end{equation*}

Similarity loss $l_s$ is based on BERTScore~\cite{bertscore} and captures the semantic similarity between $x$ and $x'$ using contextualized embeddings of tokens $\phi_g(x)=(v_1,...,v_n)$ and $\phi_g(x')=(v'_1,...,v'_n)$ produced by the reference model $g$ :
\begin{equation*}
    l_s(g, x', x) = -\sum_{i=1}^n w_i \max_j v_i^Tv'_j,
\end{equation*}
where $w_i$ is the inverse frequency of token $t_i$.

\paragraph{Quantization-Compensation Loop}
MANGO algorithm searches for a $x'$ that minimizes $\mathcal{L}$, quantizing and compensating it step by step. Algorithm~\ref{algo:mango} introduces the idea of MANGO.

In the first line, the parameters $\Theta$ of $x'$ are initialized, so that $\Theta'_{ij} = C\cdot(x_i)_j$ for some constant $C$. Each loop starts with \textbf{optimization} of $x'$ with respect to $\mathcal{L}$. Then \textbf{vector selection} is performed to select $\pi'_i$ from $x'$ which will be quantized in the current step. Given $\pi'_i$, MANGO performs \textbf{candidates selection} and selects $m$ the most promising tokens $c_1, ..., c_{m}$ to which $x'_i$ can be quantized. In the 6th line, each candidate $c_j$ is evaluated by computing $\mathcal{L}$ for a sequence $x'$ with vector $\pi'_i$ quantized to $c_j$. Finally,  $\pi'_i$ is quantized to the best $c_j$ chosen from the previous step. Quantized $\pi'_i$ will no longer be updated during optimization. MANGO repeats lines 2-7 until all vectors in $x'$ are quantized.

\begin{algorithm}[h]
\DontPrintSemicolon
 \KwData{adversarial loss $\mathcal{L}$ (eq. \ref{eq:loss})}
 \KwResult{sentence $x'$ that minimizes $\mathcal{L}$}
 initialize $x'=(\pi'_1, ..., \pi'_n)$ \;
 \While{$x'$ is not fully quantized}{
    \textbf{optimization}: optimize parameters of $x'$\;
    \textbf{vector selection}: select probabilistic vector $\pi'_i$ from $x'$ for quantization\;
    \textbf{candidates selection}: select $m$ tokens candidates from $\pi'_i$\;
    evaluate these $m$ candidates with loss $\mathcal{L}$ \; 
    quantize $\pi'_i$ to best evaluated token \;
 }
 \caption{MANGO}
 \label{algo:mango}
\end{algorithm}
\vspace{-.4cm}

\paragraph{Optimization} We optimize $x'$ with the Adam optimizer \cite{adam} which is reset after each quantization (see \Cref{app:entropy}). This allows $x'$ to rapidly change its trajectory to compensate for the degradation of $\mathcal{L}$. The initial number of optimization steps is $S$, but it decreases by a factor of 2 in each loop to reduce computational costs.

\paragraph{Vector Selection}
In line 4th, we choose vector $\pi'_i$ with the highest entropy (see \Cref{app:entropy}), because its quantization will introduce the most significant change to $x'$ 
and is likely to increase the loss value the most. Intuitively, we want such degrading quantizations to occur early in the algorithm, because the more vectors are not quantized yet, the larger capacity $x'$ has to compensate for degradation by finding another local minimum of $\mathcal{L}$.

\paragraph{Candidates Selection}
In this phase, we select $m$ tokens that can be used to quantize the probability vector $\pi'_i$ with possibly a small degradation of $\mathcal{L}$. Quantization of $\pi'_i$ with token $k$ is a step $q_k=(-(\pi'_i)_1, -(\pi'_i)_2, ..., 1 - (\pi'_i)_k, ..., -(\pi'_i)_n)$ in the $\pi'_i$ space. As $\pi'_i$ is likely to be in the proximity of its local minimum with respect to $\mathcal{L}$, we want the step $q_k$ to have (1) the lowest norm $\lVert q_k \rVert$ possible and (2) follow the direction of the local (minus) gradient. We use this intuition in the formulation of the token score $s_k$, which is a weighted mean of the probability $(\pi'_i)_k$ and the direction score $d_k$:
\begin{equation}
    s_k = \lambda_{prob}(\pi'_i)_k + (1-\lambda_{prob})d_k.
\end{equation}
Note that $(\pi'_i)_k$ is inversely proportional to $\lVert q_k \rVert$. We define $d_k$ as cosine similarity between $q_k$ and the local (minus) gradient (see \Cref{app:entropy}):
\begin{equation}
\label{eq:score}
    d_k = \frac{q_k\left( -\nabla_{\pi'_i} \mathcal{L}(x') \right)^T}{\lVert q_k \rVert \cdot \lVert \nabla_{\pi'_i} \mathcal{L}(x') \rVert}
\end{equation}
We then select $m$ tokens with the highest scores $s_k$.

%% file: content/tables/all_results.tex
\begin{table*}[t!]
\centering
\small
\begin{tabular}{clccccccc}
\toprule
Task & Method & Adv. & Adv. prob. & USE sim. & BERTScore & $\Delta$ perp. & $\Delta$ gram. & \# queries \\
\midrule

\multirow{7}{*}{\rotatebox[origin=c]{90}{\shortstack{AG News \\ (99.6)}}}
& TextFooler & 16.2 & 43.7 $\pm$ 26.0 & 0.81 $\pm$ 0.13 & 0.83 $\pm$ 0.10 & 373 $\pm$ 548 & 0.26 $\pm$ 0.69 & 334 $\pm$ 224 \\
& Bert-Attack & 20.1 & 45.7 $\pm$ 27.7 & 0.83 $\pm$ 0.11 & 0.86 $\pm$ 0.09 & 86 $\pm$ 133 & 0.06 $\pm$ 0.49 & 620 $\pm$ 472 \\
& BAE & 12.6 & 41.1 $\pm$ 24.1 & 0.78 $\pm$ 0.16 & 0.84 $\pm$ 0.11 & 157 $\pm$ 289 & 0.07 $\pm$ 0.53 & 424 $\pm$ 353 \\
\cmidrule{2-9}
& naive & 43.7 & 44.5 $\pm$ 43.1 & 0.82 $\pm$ 0.10 & 0.87 $\pm$ 0.06 & 67 $\pm$ 141 & 0.13 $\pm$ 0.62 & 102 $\pm$ 6 \\
& GBDA  & 12.9 & 13.7 $\pm$ 29.4 & 0.72 $\pm$ 0.13 & 0.80 $\pm$ 0.09 & 241 $\pm$ 382 & 0.17 $\pm$ 0.72 & 1098 $\pm$ 69 \\
& MANGO & \textbf{2.7} & 3.2 $\pm$ 15.3 & 0.78 $\pm$ 0.10 & 0.83 $\pm$ 0.06 & 30 $\pm$ 108 & 0.10 $\pm$ 0.63 & 496 $\pm$ 125 \\

\midrule
\multirow{7}{*}{\rotatebox[origin=c]{90}{\shortstack{IMDB \\ (98.2)}}}
& TextFooler & 0.6 & 34.1 $\pm$ 16.9 & 0.94 $\pm$ 0.08 & 0.93 $\pm$ 0.07 & 108 $\pm$ 214 & 01.03 $\pm$ 1.81 & 761 $\pm$ 1 000 \\
& Bert-Attack & 0.6 & 28.0 $\pm$ 18.6 & 0.96 $\pm$ 0.07 & 0.96 $\pm$ 0.05 & 19 $\pm$ 38 & 0.05 $\pm$ 0.65 & 900 $\pm$ 922 \\
& BAE & \textbf{0.2} & 29.3 $\pm$ 18.3 & 0.95 $\pm$ 0.08 & 0.95 $\pm$ 0.06 & 27 $\pm$ 59 & 0.10 $\pm$ 0.76 & 651 $\pm$ 665 \\
\cmidrule{2-9}
& naive & 30.5 & 31.1 $\pm$ 42.6 & 0.86 $\pm$ 0.09 & 0.83 $\pm$ 0.10 & 288 $\pm$ 346 & 1.56 $\pm$ 2.75 & 100 $\pm$ 13 \\
& GBDA  & 6.3 & 7.0 $\pm$ 21.3 & 0.83 $\pm$ 0.11 & 0.79 $\pm$ 0.08 & 294 $\pm$ 271 & 1.44 $\pm$ 2.22 & 1082 $\pm$ 146 \\
& MANGO & 0.3 & 0.7 $\pm$ 5.7 & 0.88 $\pm$ 0.07 & 0.83 $\pm$ 0.08 & 59 $\pm$ 73 & 0.99 $\pm$ 2.15 & 1647 $\pm$ 746 \\
\midrule

\multirow{7}{*}{\rotatebox[origin=c]{90}{\shortstack{Yelp \\ (99.9)}}}
& TextFooler & 4.5 & 31.7 $\pm$ 22.6 & 0.92 $\pm$ 0.10 & 0.93 $\pm$ 0.06 & 90 $\pm$ 192 & 0.50 $\pm$ 01.06 & 495 $\pm$ 526 \\
& Bert-Attack & \textbf{1.9} & 28.3 $\pm$ 19.1 & 0.93 $\pm$ 0.09 & 0.94 $\pm$ 0.06 & 16 $\pm$ 38 & 0.00 $\pm$ 0.55 & 665 $\pm$ 713 \\
& BAE & 2.8 & 30.5 $\pm$ 21.1 & 0.92 $\pm$ 0.11 & 0.93 $\pm$ 0.06 & 29 $\pm$ 130 & 0.06 $\pm$ 0.60 & 501 $\pm$ 525 \\
\cmidrule{2-9}
& naive & 35.1 & 35.8 $\pm$ 45.4 & 0.82 $\pm$ 0.13 & 0.84 $\pm$ 0.09 & 25 $\pm$ 84 & 0.75 $\pm$ 1.93 & 102 $\pm$ 3 \\
& GBDA  & 4.5 & 4.9 $\pm$ 18.3 & 0.79 $\pm$ 0.12 & 0.81 $\pm$ 0.06 & 5 $\pm$ 42 & 0.37 $\pm$ 1.59 & 1101 $\pm$ 35 \\
& MANGO & 8.5 & 8.9 $\pm$ 27.4 & 0.82 $\pm$ 0.12 & 0.80 $\pm$ 0.07 & -30 $\pm$ 38 & 0.34 $\pm$ 1.72 & 1128 $\pm$ 718 \\
\midrule

\multirow{7}{*}{\rotatebox[origin=c]{90}{\shortstack{MNLI premise \\ (94.7)}}}
& TextFooler & 94.7 & - & - & - & - & - & - \\
& Bert-Attack & 3.9 & 34.3 $\pm$ 23.5 & 0.93 $\pm$ 0.08 & 0.96 $\pm$ 0.04 & 30 $\pm$ 58 & 0.02 $\pm$ 0.26 & 146 $\pm$ 148 \\
& BAE & 5.0 & 34.3 $\pm$ 23.5 & 0.92 $\pm$ 0.09 & 0.95 $\pm$ 0.04 & 42 $\pm$ 107 & 0.01 $\pm$ 0.26 & 112 $\pm$ 108 \\
\cmidrule{2-9}
& naive & 31.6 & 33.9 $\pm$ 24.0 & 0.91 $\pm$ 0.07 & 0.94 $\pm$ 0.04 & 64 $\pm$ 116 & -0.01 $\pm$ 0.50 & 97 $\pm$ 23 \\
& GBDA  & 5.9 & 30.3 $\pm$ 21.9 & 0.80 $\pm$ 0.12 & 0.87 $\pm$ 0.07 & 301 $\pm$ 446 & 0.09 $\pm$ 0.67 & 1044 $\pm$ 247 \\
& MANGO & \textbf{2.4} & 31.6 $\pm$ 23.3 & 0.88 $\pm$ 0.08 & 0.91 $\pm$ 0.05 & 73 $\pm$ 123 & 0.05 $\pm$ 0.60 & 326 $\pm$ 125 \\

\midrule

\multirow{7}{*}{\rotatebox[origin=c]{90}{\shortstack{MNLI hyp. \\ (94.7)}}}
& TextFooler & 6.5 & 35.5 $\pm$ 24.2 & 0.94 $\pm$ 0.07 & 0.95 $\pm$ 0.04 & 77 $\pm$ 139 & 0.13 $\pm$ 0.39 & 77 $\pm$ 44 \\
& Bert-Attack & 2.6 & 34.3 $\pm$ 24.3 & 1.00 $\pm$ 0.01 & 0.97 $\pm$ 0.03 & 1 $\pm$ 0 & 0.00 $\pm$ 0.06 & 95 $\pm$ 62 \\
& BAE & 3.5 & 34.8 $\pm$ 24.4 & 0.95 $\pm$ 0.06 & 0.97 $\pm$ 0.03 & 29 $\pm$ 57 & 0.03 $\pm$ 0.25 & 74 $\pm$ 39 \\
\cmidrule{2-9}
& naive & 8.4 & 32.1 $\pm$ 22.7 & 0.89 $\pm$ 0.08 & 0.93 $\pm$ 0.04 & 115 $\pm$ 209 & 0.07 $\pm$ 0.36 & 97 $\pm$ 23 \\
& GBDA  & 0.6 & 27.4 $\pm$ 21.4 & 0.81 $\pm$ 0.12 & 0.89 $\pm$ 0.06 & 220 $\pm$ 454 & 0.09 $\pm$ 0.42 & 1044 $\pm$ 247 \\
& MANGO & \textbf{0.3} & 30.0 $\pm$ 22.4 & 0.89 $\pm$ 0.09 & 0.93 $\pm$ 0.04 & 85 $\pm$ 155 & 0.06 $\pm$ 0.38 & 258 $\pm$ 68 \\
\midrule

\end{tabular}
\caption{
\label{tab:all-results}
Results for black-box and white-box methods. We report: the initial training accuracy of BERT model (under Task); training accuracy under attack (Adv.); probability of ground-truth label prediction under attack (Adv. prob.); similarity between the original and perturbed text computed with USE \cite{use} (USE sim.) and with F1 
BERTScore (BERTScore); percent change in perplexity computed with GPT-2 \cite{gpt2} ($\Delta$ perpl.); increase in the number of grammar errors ($\Delta$ gram.) obtained with LanguageTool (\url{github.com/jxmorris12/language\_tool\_python}); average number of queries to a victim model (\# queries). We omit results for TextFooler on MNLI p., as it has not generated any adversarial example. We also report standard deviation for each result, except adversarial accuracy as it is simply the percent of successful attacks. Our MANGO method achieves superior results on most tasks while maintaining high semantic similarity and grammar fluency. The best results for Adv. are \textbf{bold}.
}
\end{table*}

%% file: content/experiments.tex
In this section, we evaluate MANGO on various NLP tasks and compare it to recent NLP attacks.

\paragraph{Baselines}
We compare our method with the latest white-box GBDA attack \cite{gbda}, as well as recent black-box attacks implemented in TextAttack \cite{textattack}: BERT-Attack \cite{bertattack}, BAE \cite{garg-ramakrishnan-2020-bae} and TextFooler \cite{textfooler}. To emphasize the importance of multi-step quantization, we evaluate the Naive version of MANGO that performs quantization in one step. MANGO, Naive and GBDA attacks use identical loss. All hyperparameters are listed in \cref{app:hparams}.

\paragraph{Tasks}
We attack BERT models from TextAttack fine-tuned on three text classification tasks: AG News \cite{ag_news}, Yelp Reviews \cite{ag_news}, IMDB \cite{imdb}, and MNLI task for natural language inference,  \cite{mnli}. In MNLI p., an attack is allowed to modify only the premise, and in MNLI h., only the hypothesis. For each task, we randomly select 1000 attack targets from the training set. We use a training set as it provides more challenging targets 
and is more relevant to Adversarial Training \cite{at_survey}.

\paragraph{Results}
Results can be found in \Cref{tab:all-results}. Our MANGO substantially reduces the training accuracy of the BERT model in all tasks, while maintaining a high level of semantic similarity to the original input. The attacks of MANGO are difficult (low Adv. prob., which indicates that model misclassifies an example by a large margin), fluent (low $\Delta$ perp.) and do not flaw the grammatical correctness (low $\Delta$ gram.).

In almost all settings, MANGO outperforms other attacks in terms of training accuracy, which we believe to be the fairest metric for comparing optimization-based methods with black-box ones due to inherent design biases (see \Cref{app:biases}).

MANGO surpasses the recent state-of-the-art optimization-based GBDA attack in terms of most considered metrics: in terms of Adv. acc. and BERTScore on 4/5 tasks and in terms of USE sim., $\Delta$ perpl. and $\Delta$ gram. on 5/5 tasks.

Moreover, MANGO achieves considerably better results than its Naive version, emphasizing the importance of multi-step quantization.

\paragraph{Qualitative Results}
We provide qualitative analysis of a few adversarial examples generated by BAE, GBDA, and MANGO in \Cref{app:attack-examples}.

%% file: content/ablations.tex
In this section, we evaluate three solutions from \Cref{sec:mango} that improve the core idea of multi-step quantization: 
\begin{enumerate}
    \item selection of probability vector to quantization by maximal entropy (instead of minimal entropy, which seems more natural choice),
    \item scoring token candidates by weighted mean of token probability and gradient direction score (eq. \ref{eq:score}),
    \item resetting optimizer after every quantization.
\end{enumerate}

\begin{figure}[h]
    \centering
    \includegraphics[width=0.5\textwidth]{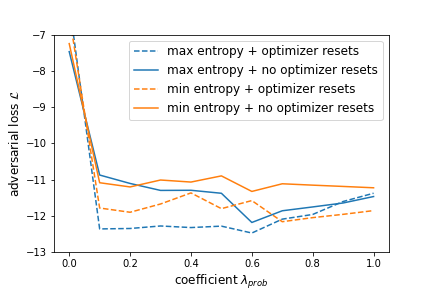}
    \caption{Final adversarial losses for different MANGO setting. "max entropy + optimizer resets" stands for a version of MANGO that selects probability vector for quantization by maximal entropy and resets optimizer after every quantization. Rest of the names follow the same pattern. We also present the influence of the coefficient $\lambda_{prob}$ used in token candidates scoring function (eq.~\ref{eq:score}). Loss values are averaged over 10 samples from IMDB dataset.}
    \label{fig:intuition}
\end{figure}

Figure~\ref{fig:intuition} compares different MANGO settings. We may observe that selection of probability vector for quantization by maximal entropy ("max entropy") is better than selection by minimal entropy ("min entropy"). Resetting the optimizer after every quantization enhances the performance for both "max entropy" and "min entropy" settings. Finally, we see that MANGO benefits from using both token's probability and gradient direction to score token candidates.

%% file: content/visualization.tex
\begin{figure}[h]
    \centering
    \includegraphics[width=0.5\textwidth]{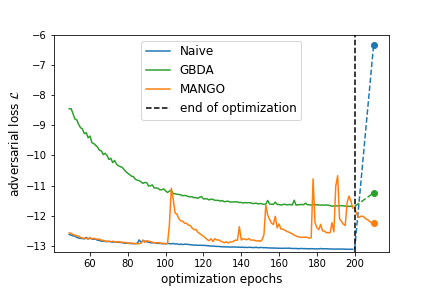}
    \caption{Adversarial loss for epochs 50-200 of optimization for Naive, GBDA and MANGO methods. The vertical dashed line shows the end of optimization. Naive and GBDA methods immediately quantize the tokens, while MANGO do it step by step. The rightmost points shows the final adversarial loss value. We observe that after optimization, MANGO continues to quantize tokens step by step and eventually reaches the best adversarial loss value. Loss values are averaged over 9 samples from IMDB dataset.}
    \label{fig:losses}
\end{figure}

To visualize the quantization gap between adversarial loss for continuous and discrete text representations, we compared adversarial losses of MANGO, GBDA and a Naive version of MANGO that does not use multi-step quantization. The comparison can be found in \Cref{fig:losses}. We observe that the Naive method converges to the lowest value loss in the optimization phase, but the value explodes after quantization. The GBDA method, which samples probability vectors that resemble discrete one-hot vectors using Gumbel-softmax \cite{gumbel}, reaches a higher minimum, but its quantization gap is much smaller than that of Naive method. Finally, in the case of MANGO, we observe sudden peaks and slow declines of loss values that correspond to the quantization-compensation loop, in which the quantization of single tokens is followed by the compensation of the quantization gap. After optimization, MANGO continues to quantize tokens step by step further decreasing the loss. MANGO obtains a significantly lower final adversarial loss than GBDA and Naive, avoiding the quantization gap.

%% file: content/conclusion.tex
We developed MANGO, a novel optimization-based attack against Transformer models that mitigates the gap between adversarial loss for continuous and discrete text representations using a quantization-compensation loop. MANGO achieves superior results on various NLP tasks, outperforming recent black-box and optimization-based attacks.


%% file: content/limitations.tex
One limitation is that the number of queries of MANGO to the attacked model depends on the length of the input sequence. Therefore, MANGO may suffer a long attack time on datasets with long sequences (like IMDB or Yelp).

Moreover, MANGO is restricted only to token replacement. The inability to insert or remove tokens can lead to reduced attack performance.

The most important limitation is the white-box nature of MANGO that excludes it from applications when the internal model's states cannot be known. To partially circumvent this limitation, we propose Gray MANGO - a version of MANGO that can be used in the loosened black-box setting, which we call gray-box setting (see \cref{app:black_mango}).

%% file: content/appendix.tex
\clearpage
\section{Hyperparameters}
\label{app:hparams}

\paragraph{MANGO}
To save computational resources during \textbf{candidates selection}, we use the dynamic number of candidates $m$. We rescale the candidate scores $s_k$ to $[0, 1]$ and take at most $M=5$ candidates whose scores differ from the best score at most by a threshold $T=0.5$: $s_k \geq \max_j s_j - T$. We use $\lambda_{prob}=0.5$ in \Cref{eq:score}.

\paragraph{White-Box Attacks}
MANGO, Naive and GBDA methods use the loss function \Cref{eq:loss} with the same parameters $\lambda_s=20$, $\lambda_f=1$, $\kappa=5$ (taken from \citet{gbda}) for all tasks, except Yelp, where they use $\lambda_s=10$. As a reference model $g$, we used the GPT-2 model downloaded from the official GBDA repository. We set $C=10$ for initialization of the adversarial sample parameters. The number of optimization epochs $S=100$ for all models and the batch size in GBDA was set to 10.

\paragraph{Black-Box Attacks} We take TextFooler, BertAttack, and BAE implementations from TextAttack \cite{textattack} along with their original parameters. For fair comparison, we set the USE similarity threshold to the lowest value (0.2) used along these methods. Following the GBDA paper, we slightly modify the BertAttack method to mitigate its problem with subtokens and extremely long time of attack.

\section{Comparison Fairness}
\label{app:biases}
When comparing the results of optimization-based (MANGO, GBDA, Naive MANGO) and black-box methods (TextFooler, Bert-Attack, BAE), we should note that black-box methods stop perturbing text as soon as they fool the model, while optimization-based attacks minimize adversarial loss (that encourage them to fool the model by some margin) for some fixed number of steps. The former improves similarity metrics (USE sim., BERTScore) and the latter highly decreases the model's prediction on ground-truth labels (Adv. prob.), increasing the difficulty of generated sample. Therefore, we believe that training accuracy under attack (Adv.) is the fairest metric to make a direct comparison between optimization-based and classic black-box methods.

\section{Attack Examples}
\label{app:attack-examples}
To draw some insights into MANGO performance, we compared examples generated by BAE, GBDA and MANGO. We chose all the sentences from AG News and MNLI hypothesis that were successfully perturbed by the three considered methods and on which the methods obtained USE cosine similarity score greater than 0.9. We then sampled two sentences from AG News and two from MNLI hypothesis tasks. To avoid cherry-picking, we fixed a seed and sampled only once. Examples can be found in \cref{tab:attack-examples-ag-news} and in \cref{tab:attack-examples-mnli}. We are careful in drawing any conclusion from the qualitative results, however, there seems to be a trend consistent with the result from \cref{tab:all-results} and our observations from \cref{app:biases}: BAE perturbs less words than GBDA and MANGO, but also achieves lower confidence of the mislassified label.

\begin{table*}[t!]
\centering
\begin{tabular}{lcc}
\toprule
Method & Prediction & Sentence \\
\midrule
\multicolumn{3}{c}{\textbf{AG News - Example no 1.}} \\
\midrule
Original & world (100\%) & \multicolumn{1}{p{10cm}}{air india trial witness said motivated by revenge ( reuters ) reuters - a desire for revenge motivated a prosecution witness to tell the air india bombing trial he had been asked to carry an mysterious suitcase on to an airliner, defense lawyers charged on wednesday.} \\
\midrule
BAE & sci/tech (61\%) & \multicolumn{1}{p{10cm}}{air india trial witness said motivated by revenge ( reuters ) \underline{website} - a desire for revenge motivated a prosecution witness to tell the air india \underline{company} \underline{s} he had been asked to carry an mysterious suitcase on to an \underline{account}, defense lawyers charged on wednesday.} \\
\midrule
GBDA & business (99\%) & \multicolumn{1}{p{10cm}}{air india trial witness said motivated by revenge - \underline{today} \underline{investigative} \underline{reuters reporting} a desire for revenge motivated criminal prosecution \underline{witnesses} to tell the air \underline{canada} \underline{strike} trial he had been asked to carry an mysterious suitcase on to an airliner, defense lawyers charged on \underline{tuesday}.} \\
\midrule
MANGO & business (100\%) & \multicolumn{1}{p{10cm}}{air indies trial witness said motivated by revenge ( reuters ) \underline{time} - a desire for revenge motivated a prosecution witness to tell the air \underline{america} \underline{arson} trial he had been asked to carry a mysterious suitcase on to an airliner, defense lawyers charged on \underline{monday}.} \\
\midrule
\multicolumn{3}{c}{\textbf{AG News - Example no 2.}} \\
\midrule
Original & business (91\%) &  \multicolumn{1}{p{10cm}}{brazil passes bankruptcy reform brazilian congress gives the green light to a long awaited overhaul of bankruptcy laws, which it hopes will reduce business and credit costs. } \\
\midrule
BAE & sci/tech (95\%) & \multicolumn{1}{p{10cm}}{brazil passes bankruptcy reform brazilian congress gives the green light to a long awaited overhaul of \underline{copyright} laws, which it hopes will reduce business and credit costs.} 
 \\
\midrule
GBDA & world (95\%) & \multicolumn{1}{p{10cm}}{ brazil passes bankruptcy reform brazilian congress gives the green light to a long awaited overhaul of \underline{privacy} laws, which it \underline{aims} will reduce \underline{tourism} and \underline{population} \underline{impacts}.} \\
\midrule
MANGO & world (99\%) & \multicolumn{1}{p{10cm}}{brazil passes \underline{golf} reform brazilian congress gives the green light to a long awaited overhaul of \underline{elections} laws, which it hopes will reduce \underline{spending} and \underline{maintenance} costs.}
 \\
\bottomrule

\end{tabular} 
\caption{\label{tab:attack-examples-ag-news} Attack examples sampled from AG News dataset.}
\end{table*}

\begin{table*}[t!]
\centering
\begin{tabular}{lcc}
\toprule
Method & Prediction & Sentence \\
\midrule
\multicolumn{3}{c}{\textbf{MNLI hypothesis - Example no 1.}} \\
\midrule
Original & contraditcion (96\%) & \multicolumn{1}{p{10cm}}{\textbf{premise}: the houses are built to a \underline{long} - standing design and are filled with embroidery, lace, and crochet work.
\; \; \; \; \; \; \; \; \; \textbf{hypothesis}: there is no embroidery in the houses.} \\
\midrule
BAE & neutral (45\%) & \multicolumn{1}{p{10cm}}{\textbf{hypothesis:} there is no \underline{fire} in the houses.} \\
\midrule
GBDA & neutral (100\%) & \multicolumn{1}{p{10cm}}{\textbf{hypothesis:}  there is \underline{liturgical} embroidery in the houses.} \\
\midrule
MANGO & neutral (99\%) & \multicolumn{1}{p{10cm}}{\textbf{hypothesis:} there is no \underline{erosion} in the \underline{ruins}.} \\
\midrule
\multicolumn{3}{c}{\textbf{MNLI hypothesis - Example no 2.}} \\
\midrule
Original & contradiction (100\%) & \multicolumn{1}{p{10cm}}{\textbf{premise}: whether the service emerges as an adaptation from primary care or as an innovation from the ed is less important than whether it can be evaluated to the satisfaction of those who make key decisions about whether it becomes part of standard practice.
\textbf{hypothesis:} key decision makers are not important to decided things.} \\
\midrule
BAE & neutral (96\%) & \multicolumn{1}{p{10cm}}{\textbf{hypothesis:}  \underline{consensus} decision makers are not important to \underline{first} \underline{things}.} \\
\midrule
GBDA &  neutral (98\%) & \multicolumn{1}{p{10cm}}{\textbf{hypothesis:} key decision makers are \underline{noted} \underline{fairchild} – \underline{emery} \underline{associates}.} \\
\midrule
MANGO &  neutral (99\%) & \multicolumn{1}{p{10cm}}{\textbf{hypothesis:} \underline{older} \underline{ahlers} are \underline{also} important \underline{in} \underline{this} \underline{regard}.} \\
\bottomrule

\end{tabular}
\caption{\label{tab:attack-examples-mnli} Attack examples sampled from MNLI hypothesis task.}
\end{table*}


\section{Gray MANGO}
\label{app:black_mango}
To circumvent the white-box nature of MANGO attack, we additionally develop Gray MANGO: a version of MANGO that can be used in the loosened black-box setting, which we call gray-box setting.

\paragraph{Gray-Box Setting} Gray MANGO is not strictly a black-box attack, as it requires the attacked model to take probability vectors and needs access to token vocabulary $V$. Transformer-based models satisfy these assumptions: they usually share the same $V$ and their embedding function $e$ can be used for both one-hot and probability vectors. However, to avoid misconception, we call this loosened black-box setting a grey-box setting.


\paragraph{Zeroth-Order Optimization} Gray MANGO is based on Zeroth-Order Optimization (ZOO) \cite{zoo}. The idea of ZOO is to approximate the gradient using only zeroth order loss values. In computer vision, \citet{zoo_images} developed a ZOO-based attack that significantly outperforms other black-box attacks. We believe that this success can be transferred to the NLP domain. \citet{dont-search} have proposed an NLP attack that uses a discrete version of ZOO, but the results were unsatisfactory. Our Gray MANGO method is the first to successfully adapt the continuous version of ZOO in NLP attacks. 

\paragraph{Formulation} The main modification with respect to MANGO is the use of the zeroth-order gradient approximation of the gradient $\nabla_{\Theta'}\mathcal{L}(x')$ \cite{zoo_many}:
\begin{equation*}
    \widetilde{\nabla}_{\Theta'} \mathcal{L}(x') = \frac{1}{K}\sum_{i=1}^{K} \frac{\mathcal{L}(\sigma(\Theta' + \mu u_i)) - \mathcal{L}(x')}{\mu}u_i,
\end{equation*}
where $u_i$ is a noise sampled from the normal distribution, $\mu$ is the scale factor and $\sigma(\Theta' + \mu u_i)$ is $x'$ with noise $\mu u_i$ added to its parameters $\Theta'$. 

As $\widetilde{\nabla}_{\Theta'}\mathcal{L}(x')$ is unstable, we set $\lambda_{prob}=1$ and use AMSGrad variant of Adam \cite{zoo_adamm} without reset after every quantization.
To reduce the high dimensionality of $x'$, which is an issue in ZOO \cite{high_dimension}, we disallow replacement of the original token with tokens that have a cosine similarity of GloVe \cite{glove} embedding lower than 0.

\paragraph{Hyperparameters} We use almost the same parameters as for MANGO (see \cref{app:hparams}), but with $\lambda_{prob} = 1$, $S=140$ and $\lambda_{s}=80$. To save computational resources, we set $S=100$ for the IMDB and Yelp datasets. Based on small grid search, we set the noise scaling parameter $\mu=0.1$.

\paragraph{Results} We evaluated the Gray MANGO method and compared it to vanilla MANGO. Results can be found in \cref{tab:black_results}. 

Gray MANGO, which is the first method to incorporate continuous ZOO in NLP attack, performs competitively with other black-box attacks in terms of training accuracy reduction, but struggles to keep adversarial examples similar to original texts. We believe that the performance of Gray MANGO may be greatly elevated by a more thorough design of ZOO components \cite{zoo_many}. This may be an interesting topic for future research.

\subfile{tables/black_mango}

%% file: content/tables/black_mango.tex
\begin{table*}[t!]
\centering
\small
\begin{tabular}{clccccccc}
\toprule
Task & Method & Adv. & Adv. prob. & USE sim. & BERTScore & $\Delta$ perp. & $\Delta$ gram. & \# queries \\
\midrule

\multirow{5}{*}{\rotatebox[origin=c]{90}{\shortstack{AG News \\ (99.6)}}}
& TextFooler & 16.2 & 43.7 $\pm$ 26.0 & 0.81 $\pm$ 0.13 & 0.83 $\pm$ 0.10 & 373 $\pm$ 548 & 0.26 $\pm$ 0.69 & 334 $\pm$ 224 \\
& Bert-Attack & 20.1 & 45.7 $\pm$ 27.7 & 0.83 $\pm$ 0.11 & 0.86 $\pm$ 0.09 & 86 $\pm$ 133 & 0.06 $\pm$ 0.49 & 620 $\pm$ 472 \\
& BAE & 12.6 & 41.1 $\pm$ 24.1 & 0.78 $\pm$ 0.16 & 0.84 $\pm$ 0.11 & 157 $\pm$ 289 & 0.07 $\pm$ 0.53 & 424 $\pm$ 353 \\
& G-MANGO & 9.7 & 11.0 $\pm$ 25.8 & 0.57 $\pm$ 0.23 & 0.67 $\pm$ 0.14 & 16k $\pm$ 47k & -0.03 $\pm$ 0.61 & 3728 $\pm$ 244 \\
& MANGO & \textbf{2.7} & 3.2 $\pm$ 15.3 & 0.78 $\pm$ 0.10 & 0.83 $\pm$ 0.06 & 30 $\pm$ 108 & 0.10 $\pm$ 0.63 & 496 $\pm$ 125 \\

\midrule
\multirow{5}{*}{\rotatebox[origin=c]{90}{\shortstack{IMDB \\ (98.2)}}}
& TextFooler & 0.6 & 34.1 $\pm$ 16.9 & 0.94 $\pm$ 0.08 & 0.93 $\pm$ 0.07 & 108 $\pm$ 214 & 01.03 $\pm$ 1.81 & 761 $\pm$ 1 000 \\
& Bert-Attack & 0.6 & 28.0 $\pm$ 18.6 & 0.96 $\pm$ 0.07 & 0.96 $\pm$ 0.05 & 19 $\pm$ 38 & 0.05 $\pm$ 0.65 & 900 $\pm$ 922 \\
& BAE & 0.2 & 29.3 $\pm$ 18.3 & 0.95 $\pm$ 0.08 & 0.95 $\pm$ 0.06 & 27 $\pm$ 59 & 0.10 $\pm$ 0.76 & 651 $\pm$ 665 \\
& G-MANGO & 8.6 & 10.8 $\pm$ 24.3 & 0.65 $\pm$ 0.21 & 0.66 $\pm$ 0.14 & 16k $\pm$ 38k & 0.19 $\pm$ 1.97 & 3142 $\pm$ 669 \\
& MANGO & 0.3 & 0.7 $\pm$ 5.7 & 0.88 $\pm$ 0.07 & 0.83 $\pm$ 0.08 & 59 $\pm$ 73 & 0.99 $\pm$ 2.15 & 1647 $\pm$ 746 \\
\midrule

\multirow{5}{*}{\rotatebox[origin=c]{90}{\shortstack{Yelp \\ (99.9)}}}
& TextFooler & 4.5 & 31.7 $\pm$ 22.6 & 0.92 $\pm$ 0.10 & 0.93 $\pm$ 0.06 & 90 $\pm$ 192 & 0.50 $\pm$ 01.06 & 495 $\pm$ 526 \\
& Bert-Attack & \textbf{1.9} & 28.3 $\pm$ 19.1 & 0.93 $\pm$ 0.09 & 0.94 $\pm$ 0.06 & 16 $\pm$ 38 & 0.00 $\pm$ 0.55 & 665 $\pm$ 713 \\
& BAE & 2.8 & 30.5 $\pm$ 21.1 & 0.92 $\pm$ 0.11 & 0.93 $\pm$ 0.06 & 29 $\pm$ 130 & 0.06 $\pm$ 0.60 & 501 $\pm$ 525 \\
& G-MANGO & 15.7 & 16.4 $\pm$ 32.1 & 0.62 $\pm$ 0.27 & 0.69 $\pm$ 0.15 & 14k $\pm$ 36k & -0.01 $\pm$ 1.68 & 2803 $\pm$ 516 \\
& MANGO & 8.5 & 8.9 $\pm$ 27.4 & 0.82 $\pm$ 0.12 & 0.80 $\pm$ 0.07 & -30 $\pm$ 38 & 0.34 $\pm$ 1.72 & 1128 $\pm$ 718 \\
\midrule

\multirow{5}{*}{\rotatebox[origin=c]{90}{\shortstack{MNLI p. \\ (94.7)}}}
& TextFooler & 94.7 & - & - & - & - & - & - \\
& Bert-Attack & 3.9 & 34.3 $\pm$ 23.5 & 0.93 $\pm$ 0.08 & 0.96 $\pm$ 0.04 & 30 $\pm$ 58 & 0.02 $\pm$ 0.26 & 146 $\pm$ 148 \\
& BAE & 5.0 & 34.3 $\pm$ 23.5 & 0.92 $\pm$ 0.09 & 0.95 $\pm$ 0.04 & 42 $\pm$ 107 & 0.01 $\pm$ 0.26 & 112 $\pm$ 108 \\
& G-MANGO & 35.1 & 33.4 $\pm$ 23.0 & 0.77 $\pm$ 0.18 & 0.84 $\pm$ 0.10 & 5876 $\pm$ 19k & -0.06 $\pm$ 0.64 & 3158 $\pm$ 761 \\
& MANGO & \textbf{2.4} & 31.6 $\pm$ 23.3 & 0.88 $\pm$ 0.08 & 0.91 $\pm$ 0.05 & 73 $\pm$ 123 & 0.05 $\pm$ 0.60 & 326 $\pm$ 125 \\

\midrule

\multirow{5}{*}{\rotatebox[origin=c]{90}{\shortstack{MNLI h. \\ (94.7)}}}
& TextFooler & 6.5 & 35.5 $\pm$ 24.2 & 0.94 $\pm$ 0.07 & 0.95 $\pm$ 0.04 & 77 $\pm$ 139 & 0.13 $\pm$ 0.39 & 77 $\pm$ 44 \\
& Bert-Attack & 2.6 & 34.3 $\pm$ 24.3 & 1.00 $\pm$ 0.01 & 0.97 $\pm$ 0.03 & 1 $\pm$ 0 & 0.00 $\pm$ 0.06 & 95 $\pm$ 62 \\
& BAE & 3.5 & 34.8 $\pm$ 24.4 & 0.95 $\pm$ 0.06 & 0.97 $\pm$ 0.03 & 29 $\pm$ 57 & 0.03 $\pm$ 0.25 & 74 $\pm$ 39 \\
& G-MANGO & 9.1 & 30.8 $\pm$ 22.4 & 0.83 $\pm$ 0.13 & 0.89 $\pm$ 0.07 & 1402 $\pm$ 3272 & 0.04 $\pm$ 0.35 & 3387 $\pm$ 807 \\
& MANGO & \textbf{0.3} & 30.0 $\pm$ 22.4 & 0.89 $\pm$ 0.09 & 0.93 $\pm$ 0.04 & 85 $\pm$ 155 & 0.06 $\pm$ 0.38 & 258 $\pm$ 68 \\
\midrule

\end{tabular}
\caption{\label{tab:black_results}
Comparison of Gray MANGO with black-box methods and vanilla MANGO. We report: the initial training accuracy of BERT model (under Task); training accuracy under attack (Adv.); probability of ground-truth label prediction under attack (Adv. prob.); similarity between the original and perturbed text computed with USE \cite{use} (USE sim.) and with F1 BERTScore (BERTScore); percent change in perplexity computed with GPT-2 \cite{gpt2} ($\Delta$ perpl.); increase in the number of grammar errors ($\Delta$ gram.) obtained with LanguageTool (\url{github.com/jxmorris12/language\_tool\_python}); average number of queries to a victim model (\# queries). We omit results for TextFooler on MNLI p., as it has not generated any adversarial example. We also report standard deviation for each result, except adversarial accuracy as it is simply the percent of successful attacks. The best results for Adv. are \textbf{bold}.
}
\end{table*}